\PassOptionsToPackage{dvipsnames,table}{xcolor}%
\documentclass[10pt,twocolumn,letterpaper]{article}

\usepackage{iccv}              %

\makeatletter
\renewcommand\paragraph{\@startsection{paragraph}{4}{\z@}{-0.8mm}{-2.8mm}{\bf\itshape\@adddotafter}}
\makeatother

\usepackage{array}

\newcommand{\mparagraph}[1]{\vspace{1mm}\noindent{\textbf{#1.}\hspace{1mm}}}

\usepackage{import} %
\graphicspath{{figs/}}

\usepackage[ruled,noend]{algorithm2e} %

\SetAlFnt{\small}
\SetAlCapFnt{\small}
\SetAlCapNameFnt{\small}
\SetAlCapHSkip{0pt}
\IncMargin{-\parindent}
\usepackage{ifpdf}
\usepackage{graphicx}
\ifpdf
  
\else
\fi
\usepackage{color}
\definecolor{cyan}{cmyk}{1,0,0,0}
\definecolor{darkgreen}{rgb}{0,0.5,0}
\definecolor{orange}{rgb}{1,0.5,0}
\definecolor{magenta}{cmyk}{0,1,0,0}
\definecolor{darkyellow}{cmyk}{0,0,0.75,0}
\definecolor{gray}{rgb}{0.8,0.8,0.8}

\usepackage{amsmath} %
\usepackage{amssymb} %

\usepackage[toc,page]{appendix} %
\usepackage{wrapfig} %
\usepackage{comment}
\usepackage{bm}
\usepackage{cuted} %
\usepackage{lipsum} %
\usepackage{mathtools} %
\usepackage{algorithmicx}
\usepackage{algpseudocode}
\usepackage{nicefrac}

\makeatletter
\let\ESO@isMEMOIR\relax
\let\AtPageUpperLeft\relax

\let\AtTextUpperLeft\relax

\let\ESO@HookI\relax
\let\ESO@HookII\relax
\let\ESO@HookIII\relax
\let\AddToShipoutPicture\relax

\let\ESO@yoffsetI\relax
\let\ESO@yoffsetII\relax
\makeatother

\usepackage{pdfpages}

\usepackage{gensymb}
\usepackage{float}
\usepackage{soul}
\usepackage{array}
\usepackage{multirow}
\usepackage{hhline}
\usepackage{setspace}
\usepackage{nicefrac}

\algdef{SE}[DOWHILE]{Do}{doWhile}{\algorithmicdo}[1]{\algorithmicwhile\ #1}%

\makeatletter
\renewcommand{\ALG@beginalgorithmic}{\small}
\makeatother

\newcommand{\DELETE}[1]{} %
\newcommand{\IGNORE}[1]{}
\usepackage{datenumber}
\usepackage{calc}
\usepackage[mmddyyyy]{datetime}

\newcounter{datetoday}
\newcounter{diffyears}
\newcounter{diffmonths}
\newcounter{diffdays}

\newcommand{\difftoday}[3]{%
      \setmydatenumber{datetoday}{\the\year}{\the\month}{\the\day}%
      \setmydatenumber{diffdays}{#1}{#2}{#3}%
      \addtocounter{diffdays}{-\thedatetoday}%
      \ifnum\value{diffdays}>0
        \def\diffbefore{}%
        \def\diffafter{left}%
      \else
        \def\diffbefore{}%
        \def\diffafter{ago}%
        \setcounter{diffdays}{-\value{diffdays}}%
      \fi
      \setcounter{diffyears}{\value{diffdays}/365}%
      \setcounter{diffdays}{\value{diffdays}-365*\value{diffyears}}%
      \setcounter{diffmonths}{\value{diffdays}/30}%
      \setcounter{diffdays}{\value{diffdays}-30*\value{diffmonths}}%
      \diffbefore
      \ifnum\value{diffyears}=0
      \else
        \ifnum\value{diffyears}>1
            \thediffyears\space years,
        \else
            \thediffyears\space year,
        \fi
      \fi
      \ifnum\value{diffmonths}=0
      \else
        \ifnum\value{diffmonths}>1
            \thediffmonths\space months
        \else
            \thediffmonths\space month
        \fi
      \fi
      \ifnum\value{diffdays}=0
      \else
        \ifnum\value{diffdays}>1
            \thediffdays\space days
        \else
            \thediffdays\space day
        \fi
      \fi
      \diffafter
}

\definecolor{iccvblue}{rgb}{0.21,0.49,0.74}
\usepackage[pagebackref,breaklinks,colorlinks,allcolors=iccvblue]{hyperref}

\def\paperID{02884} %
\def\confName{ICCV}
\def\confYear{2025}

\title{Splat-based 3D Scene Reconstruction with Extreme Motion-blur}

\author{Hyeonjoong Jang$^\dagger$$^\ast$ ~ ~ Dongyoung Choi$^\dagger$ ~ ~ Donggun Kim$^\dagger$ ~ ~ Woohyun Kang$^\dagger$ ~ ~ Min H. Kim$^\dagger$$^\ast$\\
$^\dagger$\,KAIST ~ ~ ~ ~ ~ ~ ~ $^\ast$\,HYPERGRAM\\
{\tt\small \{hjjang; dychoi; dgkim; whkang; minhkim\}@vclab.kaist.ac.kr}
}

\begin{document}

\twocolumn[{%
\maketitle
    \centering
    \vspace{-4mm}    
    \captionsetup{type=figure}
    \includegraphics[width=\linewidth]{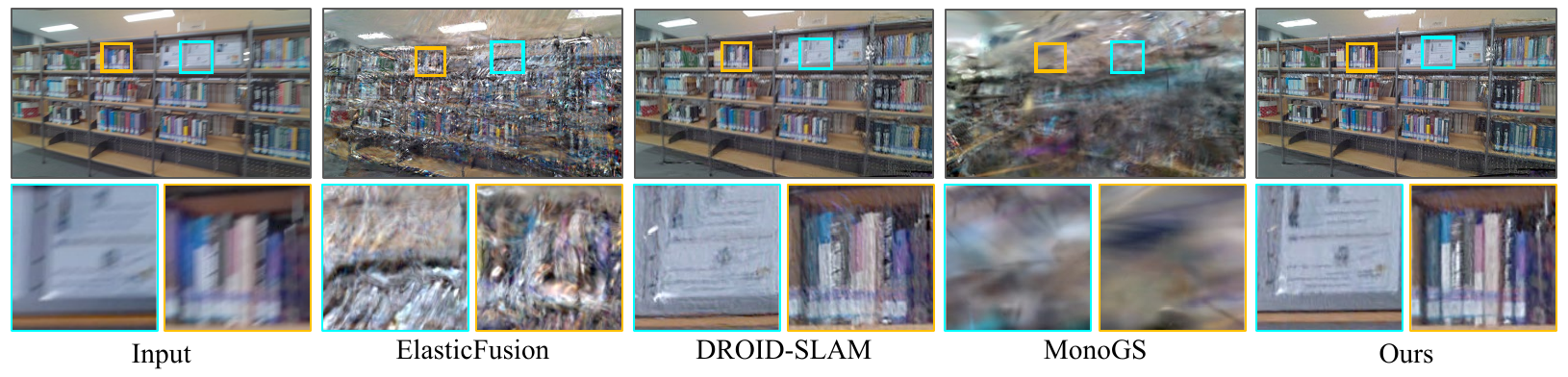}
    \vspace{-8mm}
    \captionof{figure}{
        We propose a robust 3D scene reconstruction method from RGB-D input that effectively addresses extreme motion blur. Our approach achieves accurate camera poses and dense point clouds, producing clearer, deblurred scenes compared to existing methods.
    }
    \vspace{+3mm}
    \label{fig:teaser}
}]

\begin{abstract}

\noindent 
We propose a splat-based 3D scene reconstruction method from RGB-D input that effectively handles extreme motion blur, a frequent challenge in low-light environments. 
Under dim illumination, RGB frames often suffer from severe motion blur due to extended exposure times, causing traditional camera pose estimation methods, such as COLMAP, to fail. 
This results in inaccurate camera pose and blurry color input, compromising the quality of 3D reconstructions. 
Although recent 3D reconstruction techniques like Neural Radiance Fields and Gaussian Splatting have demonstrated impressive results, they rely on accurate camera trajectory estimation, which becomes challenging under fast motion or poor lighting conditions.
Furthermore, rapid camera movement and the limited field of view of depth sensors reduce point cloud overlap, limiting the effectiveness of pose estimation with the ICP algorithm.
To address these issues, we introduce a method that combines camera pose estimation and image deblurring using a Gaussian Splatting framework, leveraging both 3D Gaussian splats and depth inputs for enhanced scene representation. 
Our method first aligns consecutive RGB-D frames through optical flow and ICP, then refines camera poses and 3D geometry by adjusting Gaussian positions for optimal depth alignment. 
To handle motion blur, we model camera movement during exposure and deblur images by comparing the input with a series of sharp, rendered frames.
Experiments on a new RGB-D dataset with extreme motion blur show that our method outperforms existing approaches, enabling high-quality reconstructions even in challenging conditions. 
This approach has broad implications for 3D mapping applications in robotics, autonomous navigation, and augmented reality. 
Both code and dataset are publicly available on \url{https://github.com/KAIST-VCLAB/gs-extreme-motion-blur}.
\end{abstract}

\section{Introduction}
\label{sec:intro}

High-quality 3D scene reconstruction is one of the most important and challenging applications in computer vision. 
The accuracy of 3D reconstruction hinges on the quality of essential components, such as camera poses, RGB images, and depth maps. 
These elements are interconnected and influence each other; a failure in one component can adversely impact the others. 
For example, blurry RGB images or noisy depth maps can significantly impair camera pose estimation, which, in turn, degrades overall reconstruction quality. 
This challenge is particularly pronounced when input data is captured in low-light conditions or during rapid camera movement—common scenarios in everyday, casual capture. 
Such conditions result in degraded color and depth frames, leading to poorly estimated camera poses and reconstructions with blurry textures and distorted or smoothed geometry. 
Therefore, achieving high-quality 3D reconstruction requires clear RGB images and accurate depth maps.

The interdependence of these input components suggests an opportunity for compensating or restoring degraded elements. 
However, this process presents a chicken-and-egg problem, complicating the reconstruction further. 
For instance, accurate camera pose estimation requires sharp images, while deblurring motion-blurred images depend on reliable camera pose information. 
Although numerous methods~\cite{tao2018srndeblur, chen2022simple} exist for recovering sharp images from motion-blurred ones, they often rely on supervised learning models trained on specific datasets. 
Consequently, their deblurring performance can degrade significantly when faced with new cameras or motion blur scenarios that differ from the training data.

Recent works~\cite{zhao2024badgaussians, wang2023badnerf, Chen_deblurgs2024, seiskari2024gaussian} have made advances in addressing these challenges by optimizing camera trajectories during exposure while simultaneously learning sharp RGB colors. 
This enables simultaneous camera pose estimation and image deblurring. 
However, these methods require an initial camera pose and, when using Gaussian Splatting~\cite{kerbl20233d}, also need a sparse point cloud as input. 
This requirement limits their applicability, especially when severe motion blur prevents structure-from-motion (SfM) methods like COLMAP~\cite{schoenberger2016sfm} from functioning effectively, making it impossible to start the optimization process.

To overcome these limitations, we introduce a robust approach for 3D scene reconstruction that effectively compensates for severe motion blur and addresses the challenges of camera pose estimation without relying on a precise initial pose. 
Our method leverages RGB-D inputs and incorporates both optical flow and depth information to align camera poses accurately, even in the presence of challenging motion blur and lighting conditions. 
We first perform a global alignment between consecutive frames using optical flow and the ICP algorithm, which enables effective local alignment of point clouds generated from the depth maps.

After this initial alignment, we refine the camera poses and 3D geometry by integrating them into a Gaussian Splatting pipeline. 
This approach allows us to initialize dense 3D Gaussians from depth maps, which we scale to ensure a detailed representation of the scene geometry. 
Our refinement process iteratively adjusts both camera poses and the positions of 3D Gaussians, achieving global consistency by minimizing a depth alignment loss that compares rendered depth maps with input depth measurements. 
Through this adjustment, we eliminate loop-closure artifacts and reduce the drift that often accumulates in traditional SLAM-based methods.

For scenes with significant motion blur, we further optimize the deblurring process by modeling the camera poses at the start and end of each frame’s exposure. 
This temporal modeling allows us to simulate the effects of motion during the exposure period, which we incorporate into our Gaussian Splatting framework. 
We minimize the difference between the observed motion-blurred image and the average of a set of sharp images rendered from multiple viewpoints along the exposure trajectory, resulting in more accurate, geometrically consistent deblurring. 
Our deblur loss function combines image alignment, structural similarity, and depth consistency, ensuring that the final reconstructions retain sharpness and fidelity to the original scene structure.

Through extensive evaluation on a newly constructed RGB-D dataset featuring extreme motion blur, we demonstrate that our approach significantly outperforms existing methods in both accuracy and robustness. 
By addressing the key challenges of pose estimation, depth alignment, and image deblurring, our method provides a versatile and effective solution for high-quality 3D reconstruction under real-world capture conditions, such as low light and fast motion. 
The proposed method holds promising applications in areas requiring reliable 3D mapping and reconstruction, including robotics, autonomous navigation, and augmented reality. 
We will make both our dataset and implementation publicly available to foster further research and development in this field.

\section{Related Work}
\label{sec:related-work}

\begin{figure*}
    \vspace{-7mm}
    \centering
   \includegraphics[width=0.85\linewidth]{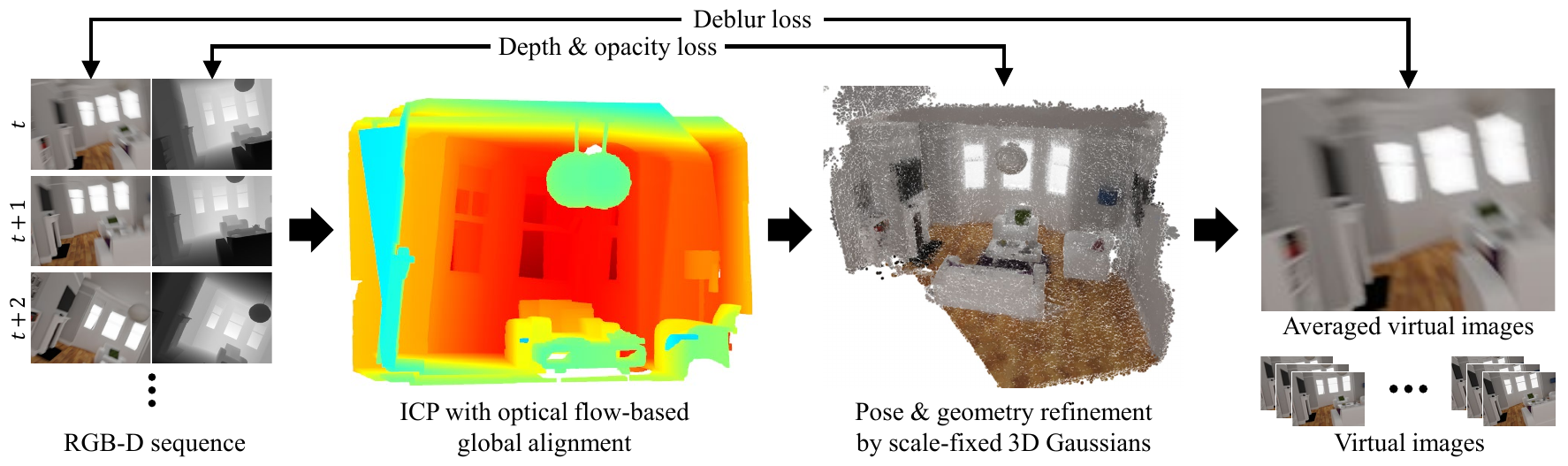}
    \vspace{-2mm}
    \caption{\label{fig:overview}
    Overview of our method pipeline. We begin with global alignment by comparing optical flow and reprojected pixel differences between adjacent RGB-D image pairs, followed by local estimation of relative camera poses using ICP. 
    Each depth map and camera pose is then used to initialize 3D Gaussians. 
    By fixing the scale of the Gaussians and optimizing with depth and opacity losses, 
    we achieve global refinement of camera poses and geometry. 
    Finally, we deblur the scene texture by minimizing the deblur loss between the input image and averaged virtual images.
    }
    \vspace{-3mm}
  \end{figure*}

\mparagraph{RGB-D scene reconstruction}
A popular approach to 3D geometry reconstruction defines a 3D voxel grid and constructs a signed distance function (SDF) volume by accumulating depth estimates captured by sensors. 
KinectFusion~\cite{newcombe2011kinectfusion} uses a Kinect camera to capture RGB-D images, then estimates relative camera poses between adjacent frames with the ICP algorithm. 
Next, it constructs and updates a truncated signed distance function (TSDF) volume, enabling real-time 3D reconstruction of objects. 
Niessner et al.~\cite{niessner2013real} improve memory efficiency by combining hashing algorithms to store and update SDF values only where needed, thus reducing memory requirements.

Several methods~\cite{genova2019deep,kahler2015hierarchical,marniok2017efficient,mccormac2017semanticfusion} adopt memory-efficient, hierarchical data structures for depth fusion. 
BundleFusion~\cite{dai2017bundlefusion} simultaneously refines camera poses and geometry using bundle adjustment, improving reconstruction accuracy and addressing the loop closure problem caused by accumulated camera pose and depth estimation errors. 
ElasticFusion~\cite{whelan2016elasticfusion} builds a map with surfels—point primitives containing position, normal, and radius—and updates the map through a deformation graph to minimize error and perform loop closure. 
Other methods~\cite{keller2013real,stuckler2014multi, schops2019surfelmeshing} directly fuse depth estimates into point clouds, optimizing memory use. 
However, these methods generally require slow camera motion to capture sharp images and accurately estimate camera transformations.

Simultaneous localization and mapping (SLAM) methods~\cite{kerl2013dense, whelan2015real, mur2017orb, schops2019bad, campos2021orb} track camera motion and construct a map in real-time by extracting visual features from input images to establish 3D point correspondences. 
CodeSLAM~\cite{bloesch2018codeslam} and DeepTAM~\cite{zhou2018deeptam} use neural networks to combine depth maps with color images, while Azinovi\'{c} et al.~\cite{Azinovic_2022_CVPR} train multi-layer perceptrons (MLP) to learn SDF and color values for scene reconstruction.
Their approach also optimizes camera transformation correction variables during training to refine poses. 
However, these methods struggle with input images that contain motion blur.
Nice-SLAM~\cite{zhu2022nice}, Point-SLAM~\cite{sandstrom2023point}, and DROID-SLAM~\cite{teed2021droid} use neural networks to track camera poses. SplaTAM~\cite{keetha2024splatam} and MonoGS~\cite{matsuki2024gaussian} introduced SLAM solutions based on 3D Gaussian Splatting. However, these methods are not designed to handle motion blur.

\mparagraph{Image deblurring}
Motion blur occurs when camera motion during the exposure time changes the projected pixel coordinates of rays on the camera sensor. 
Early studies~\cite{fergus2006removing, shan2008high, krishnan2009fast} tackle this issue by developing kernels to restore blurry images. 
Convolutional neural networks (CNNs) demonstrate powerful performance in learning deblur kernels~\cite{sun2015learning, schuler2015learning}, and deep learning techniques such as ResNet with skip connections and multi-scale networks further improved restoration quality~\cite{su2017deep, nah2017deep}. 
SRN-DeblurNet~\cite{tao2018srndeblur} adopt recurrent networks~\cite{graves2012long, shi2015convolutional} and combine them with an encoder-decoder network structure, while NAFNet~\cite{chen2022simple} simplify the network structure, extracting only essential components and proposing a nonlinear activation-free network. 
However, since these methods rely on curated datasets like the GoPro dataset~\cite{nah2017deep}, they struggle to effectively deblur images with extreme motion blur or those captured on different cameras.

There are NeRF-based approaches~\cite{mildenhall2021nerf} that tackle motion blur in input images.
Deblur-NeRF~\cite{ma2022deblur} and DP-NeRF~\cite{lee2023dp} use a set of motion-blurred images along with camera poses estimated by COLMAP as input to restore sharp images through NeRF optimization. 
BAD-NeRF~\cite{wang2023badnerf} enhances this by jointly optimizing virtual camera poses and radiance fields during exposure.
The latest approaches utilize Gaussian Splatting~\cite{kerbl20233d}, which offers faster training times than NeRF by rendering a set of 3D Gaussians with a dedicated rasterizer rather than optimizing a neural network. 
Gaussian Splatting-based deblurring methods~\cite{zhao2024badgaussians, seiskari2024gaussian, Chen_deblurgs2024} optimize the virtual camera trajectory and minimize differences between input blurry images and the average of rendered images at each virtual camera position.

All of these methods rely on initial camera poses, typically obtained from COLMAP, and for Gaussian Splatting-based methods, a sparse point cloud as well. 
However, structure-from-motion methods like COLMAP often fail with blurry inputs—such as images captured by fast-moving cameras or in low-light conditions that require long exposure times. 
Even with dense depth inputs, ICP requires acceptable global alignment, and recent point cloud matching methods~\cite{gojcic2019perfect, li2022lepard} cannot accurately determine relative camera transformations between two point clouds if they consist mainly of planar structures.

To address these issues, we propose an effective camera pose estimation algorithm tailored for RGB-D image sequences that jointly refines both the camera trajectory and the reconstructed 3D point cloud, resulting in improved sharpness in 3D reconstruction.

\section{Method}
\label{sec:method}
\mparagraph{Overview} 
In datasets with significant motion blur, using COLMAP to estimate camera poses becomes impractical. 
Additionally, rapid camera motion and the limited field of view of depth sensors further complicate this task, 
making ICP registration less effective. 
To address these challenges, we first perform global alignment using both RGB and depth images, 
which establishes sufficient initial alignment for the ICP algorithm to operate effectively. 
After achieving this global alignment, we apply ICP again to refine and accurately estimate the camera poses.
However, estimating camera poses by analyzing only two consecutive frames in a sequence can result in accumulated errors,
leading to drift over time. 
To resolve this issue, it is crucial to adjust the 3D geometry and align the camera poses across the entire sequence 
so that depth maps and camera views from all perspectives remain well-aligned. 
We accomplish this through bundle adjustment, optimizing the positions of 3D Gaussians and camera poses within the Gaussian Splatting pipeline.
Finally, to account for motion blur, we approximate the camera trajectory over time and optimize the movement during exposure, enabling effective deblurring.
See Figure~\ref{fig:overview} for an overview.

\subsection{Local Pose Estimation}
\label{sec:initial-pose}
To estimate relative camera transformations, we select consecutive frames from the input RGB-D image sequence. 
Our approach begins with global alignment using optical flow, followed by local alignment of the two-point clouds using the ICP algorithm. 
For each consecutive RGB image and depth map pair at times $t$ and $t+1$, 
we optimize the transformation $\bm{\xi} \in \mathfrak{se}(3)$ between the frames. 
Here, ${\bm{\xi} _{t \to t + 1}} = {[{\mathbf{x}^{\intercal}},{\bm{\omega} ^{\intercal}}]^{\intercal}}$, where $\mathbf{x}$ represents the translation vector between the two camera origins, 
and $\bm{\omega}$ is the rotation element of $\mathfrak{se}(3)$. 
We ultimately calculate a camera-to-world transformation matrix $\bm{\xi}_t$ for each timestamp $t$.

To achieve this, we first compute the difference between the estimated optical flow and the reprojected pixel from the camera at time $t$ to $t+1$ using the depth estimate for each pixel in the depth map ${D}_{t}$ and the intrinsic matrix $\mathbf{K}$. A pixel $\mathbf{p}_i$ in the depth map is backprojected to 3D space as:
  ${\mathbf{P}_i} = {\pi ^{ - 1}}(\mathbf{K},\mathbf{p}_i,{D}_t(\mathbf{p}_i))$.
We then project this 3D point to the camera at $t+1$:
  $\mathbf{p_j} = \pi (\mathbf{K},{\text{exp}} (\bm{\xi} _{t \rightarrow t + 1}^ \wedge )\mathbf{P}_i)$,
where $\pi$ and $\pi^{-1}$ represent the projection and backprojection operators, respectively. 
The pixel movement from $\mathbf{p}_i$ to $\mathbf{p}_j$ is then compared with the estimated optical flow $F_{t \to t+1}(\mathbf{p}_i)$ to compute the optical flow loss:
\begin{equation}
  {\mathcal{L}_F} = \sum_i{M_t(\mathbf{p}_i)M_{t+1}(\mathbf{p}_j){\left\| {{F_{t \to t + 1}}({\mathbf{p}_i}) - ({\mathbf{p}_j} - \mathbf{p}_i)} \right\|_2}}.\nonumber
\end{equation}

\begin{figure}[tp]
\centering
  \vspace{-3mm}
  \includegraphics[width=0.75\linewidth]{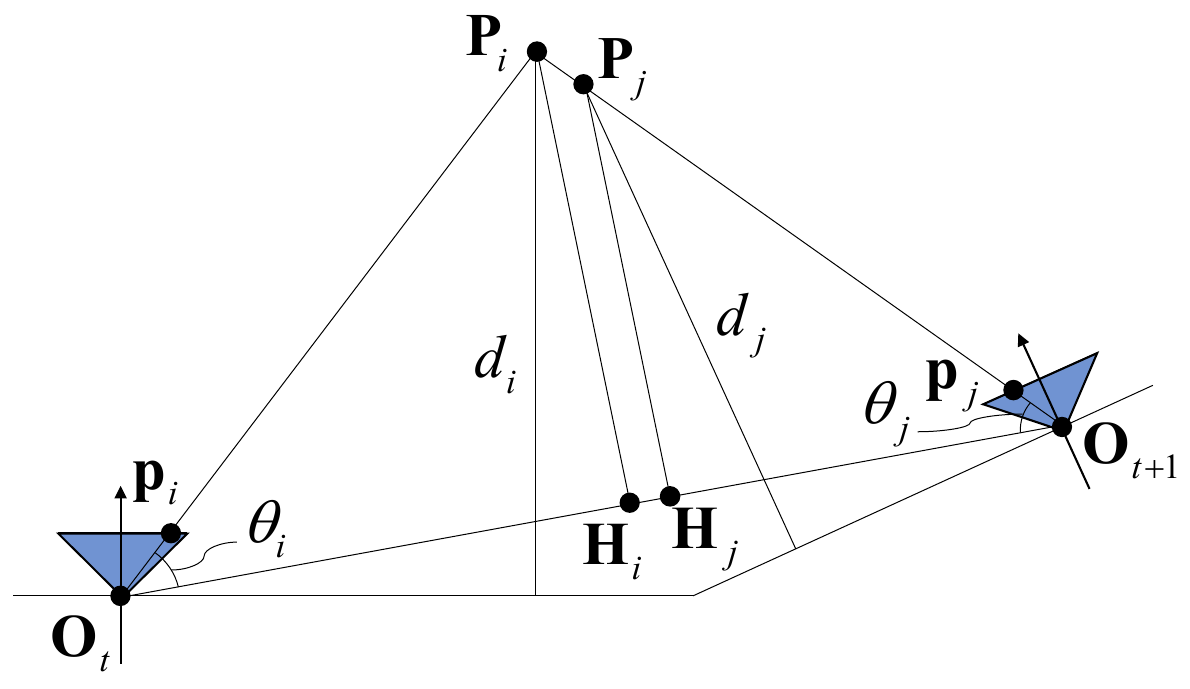}
  \vspace{-3mm}
  \caption{\label{fig:pose_initialization}
  The illustration of our pose initialization loss.
  When $\mathbf{p}_i$ is reprojected to $\mathbf{p}_j$ using the depth value, it is considered geometrically consistent if $\overline {{\mathbf{P}_i\mathbf{H}_i}}$ equals $\overline {{\mathbf{P}_j\mathbf{H}_j}}$.}
  \vspace{-4mm}
\end{figure}

In this formulation, $M_t$ is a mask indicating valid depth values in ${D}_t$ at pixel $\mathbf{p}_i$. The mask $M_t$ dynamically updates during optimization, as $\mathbf{p}_j$ depends on the transformation $\bm{\xi}$:
  $M_t(\mathbf{p}_i) = {D}_t(\mathbf{p}_i) > 0$.

We employ a state-of-the-art pretrained optical flow estimation model~\cite{xu2023unifying} designed hierarchically to produce reliable optical flow even from two blurry input images. Next, we calculate the photometric loss between color values at pixel $\mathbf{p}_i$ in image ${I}_t$ and pixel $\mathbf{p}_j$ in image ${I}_{t+1}$:
\begin{equation}
  {\mathcal{L}_I} = \sum_i {M_t(\mathbf{p}_i)M_{t+1}(\mathbf{p}_j){\left\| { {I}_t(\mathbf{p}_i) - {I}_{t+1}(\mathbf{p}_j) } \right\|_1}}.\nonumber
\end{equation}

To ensure geometric consistency, we introduce a depth consistency loss by using the two depth maps captured by the depth camera to optimize relative poses. Specifically, as illustrated in Figure~\ref{fig:pose_initialization}, we compute and compare the vertical components $\overline {{\mathbf{P}_i\mathbf{H}_i}}$ and $\overline {{\mathbf{P}_j\mathbf{H}_j}}$ based on the baseline vector $\overline {{\mathbf{O}_t\mathbf{O}_{t+1}}}$ between the two cameras. We calculate $\overline {{\mathbf{P}_i\mathbf{H}_i}}$ as follows:
\begin{equation}
  \overline {{\mathbf{P}_i\mathbf{H}_i}} = {{{\left\| {\mathbf{P}_i} \right\|}_2}\sin \left({{\cos }^{ - 1}}\left(\frac{{\mathbf{P}_i \cdot \mathbf{x}}}{{{{\left\| {\mathbf{P}_i} \right\|}_2}{{\left\| \mathbf{x} \right\|}_2}}}\right)\right)}.\nonumber
\end{equation}
Similarly, we compute $\overline {{\mathbf{P}_j\mathbf{H}_j}}$ as:
\begin{equation}
	\resizebox{\linewidth}{!}{
	\mbox{\fontsize{10}{12}\selectfont $
  \overline {{\mathbf{P}_j\mathbf{H}_j}} = {{{\left\| {\mathbf{P}_j} \right\|}_2}\sin \left(  {{\cos }^{ - 1}}\left(\frac{{\mathbf{P}_j \cdot ( - {\text{exp}} {{(\bm{\omega} )}^{\intercal}}\mathbf{x})}}{{{{\left\| {\mathbf{P}_j} \right\|}_2}{{\left\| { - {\text{exp}} {{(\bm{\omega} )}^{\intercal}}\mathbf{x}} \right\|}_2}}}\right) \right) }.
$ } } %
\nonumber
\end{equation}
This depth consistency loss is calculated as follows:
\begin{equation}
  \mathcal{L}_{C} = \sum_i {M_t(\mathbf{p}_i)M_{t+1}(\mathbf{p}_j){\left\| \overline {{\mathbf{P}_i\mathbf{H}_i}} - \overline {{\mathbf{P}_j\mathbf{H}_j}}  \right\|}},
\end{equation}
where ${\text{exp}}()$ is the exponential mapping operator in Lie algebra, which helps maintain alignment. 
Our final loss function combines these terms with weighted coefficients:
\begin{equation}
  \mathcal{L}_\text{pose}= {\mathcal{L}_F} + {\lambda _I}{\mathcal{L}_I} + {\lambda _C}{\mathcal{L}_C}.
\end{equation}

We start optimization with ${\lambda _I} = 0$ and ${\lambda _C} = 0$, allowing only the flow loss ${\mathcal{L}_F}$ to guide the relative pose alignment with global optical flow. After initial convergence, we increase weights to ${\lambda _I} = 75$ and ${\lambda _C} = 25$ and continue optimization to a second convergence.

Finally, we apply a point-to-plane ICP algorithm on the two globally aligned point clouds to estimate camera poses accurately. Since optical flow estimation between two blurry images introduces some errors, this final camera pose estimation relies exclusively on depth maps.

\subsection{Global Pose Estimation/Geometry Refinement}
\label{sec:global_refinement}

\begin{figure}
  \vspace{-3mm}
  \includegraphics[width=\linewidth]{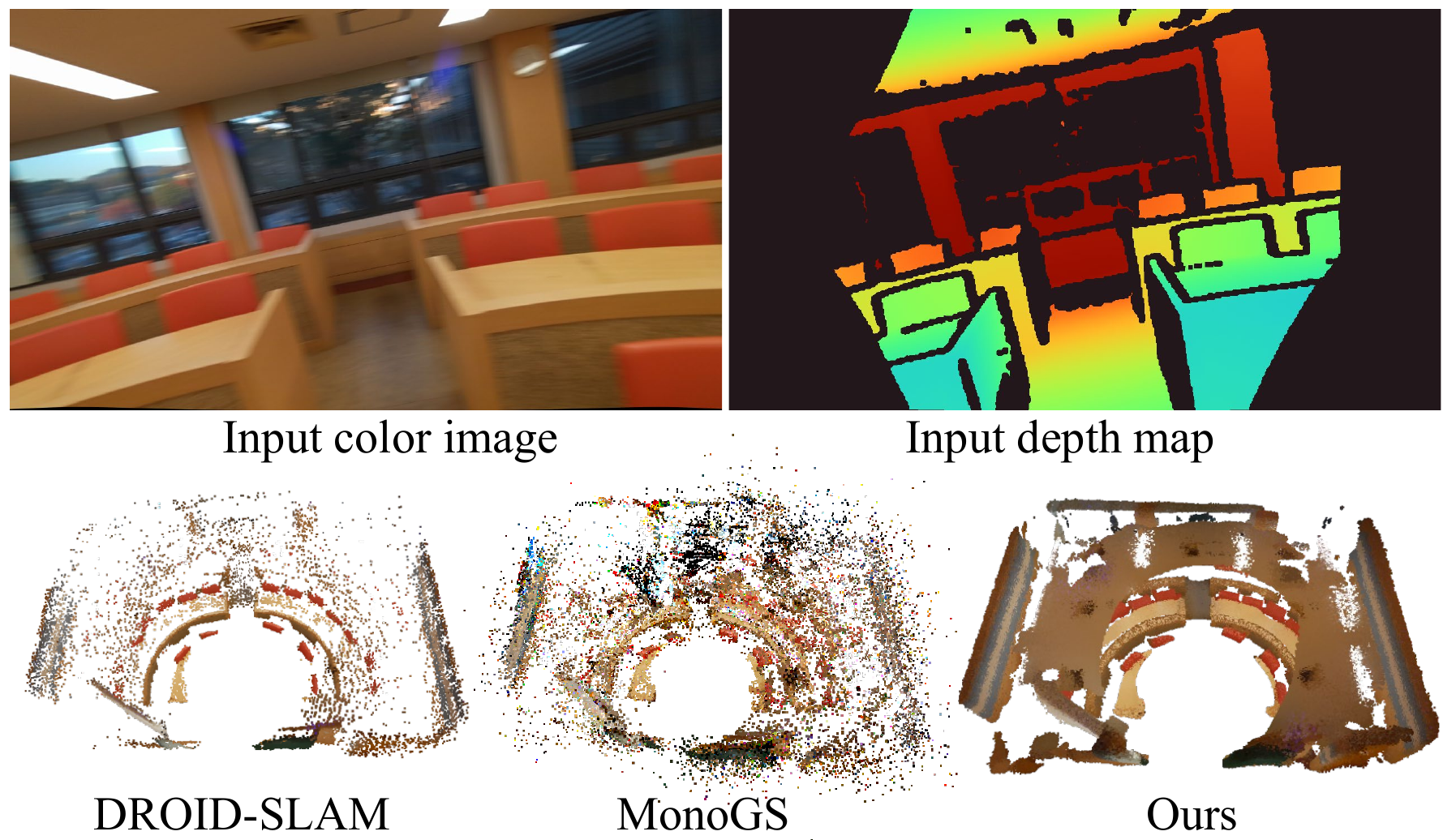}
  \vspace{-6mm}
  \caption{\label{fig:pcd}
Reconstructed initial point clouds from the RGB-D image sequence. Our method achieves accurate camera poses and generates a dense, high-quality point cloud.}
  \vspace{-3mm}
\end{figure}

Starting from the initial camera poses and point clouds obtained in the previous step, we merge them into a global point cloud within the world coordinate system (Figure~\ref{fig:pcd}). 
To manage memory efficiently, we downsample the point cloud with a specified voxel size $s$. However, due to accumulated errors in the initial camera poses, these poses may not align well with the global point cloud, leading to seam artifacts at image boundaries and a drifting phenomenon that creates loop-closure issues. To resolve this, we refine both the camera poses and 3D point cloud positions simultaneously within the Gaussian Splatting pipeline, ensuring global and geometric consistency.

We initialize the Gaussian Splatting pipeline by loading the point cloud along with the estimated camera poses. To capture every depth value per pixel from the dense depth maps, we initialize 3D Gaussians, fixing their sizes and setting their scale to $s/2$. This dense arrangement of 3D Gaussians provides a detailed geometric representation. We then train only the camera poses, 3D positions, and opacities of the Gaussians by comparing a rendered depth map ${\widetilde D}$ with the input depth map ${D}$.

During training, we randomly select camera viewpoints from among the trainable camera pose variables to render each depth map. For a depth value ${\widetilde D}_t(\mathbf{p}_i)$ at pixel $\mathbf{p}_i$ in the rendered map, we calculate the depth loss as:
\begin{equation}
{\mathcal{L}_{D}} = {\sum\limits_i {{M_t}(\mathbf{p}_i){{\left\| { {{D_t}}(\mathbf{p}_i) - {\widetilde D}_{t}(\mathbf{p}_i)} \right\|}_1}} }.
\end{equation}
We also initialize each 3D Gaussian’s opacity $o_j$ to 0.5, optimizing them to values of 0 or 1:
\begin{equation}
\label{eq:opacity_loss}
{\mathcal{L}_\text{opacity}} = \sum\limits_j {o_j^2{{(1 - {o_j})}^2}}.
\end{equation}

\begin{figure}
  \vspace{-3mm}
  \includegraphics[width=\linewidth]{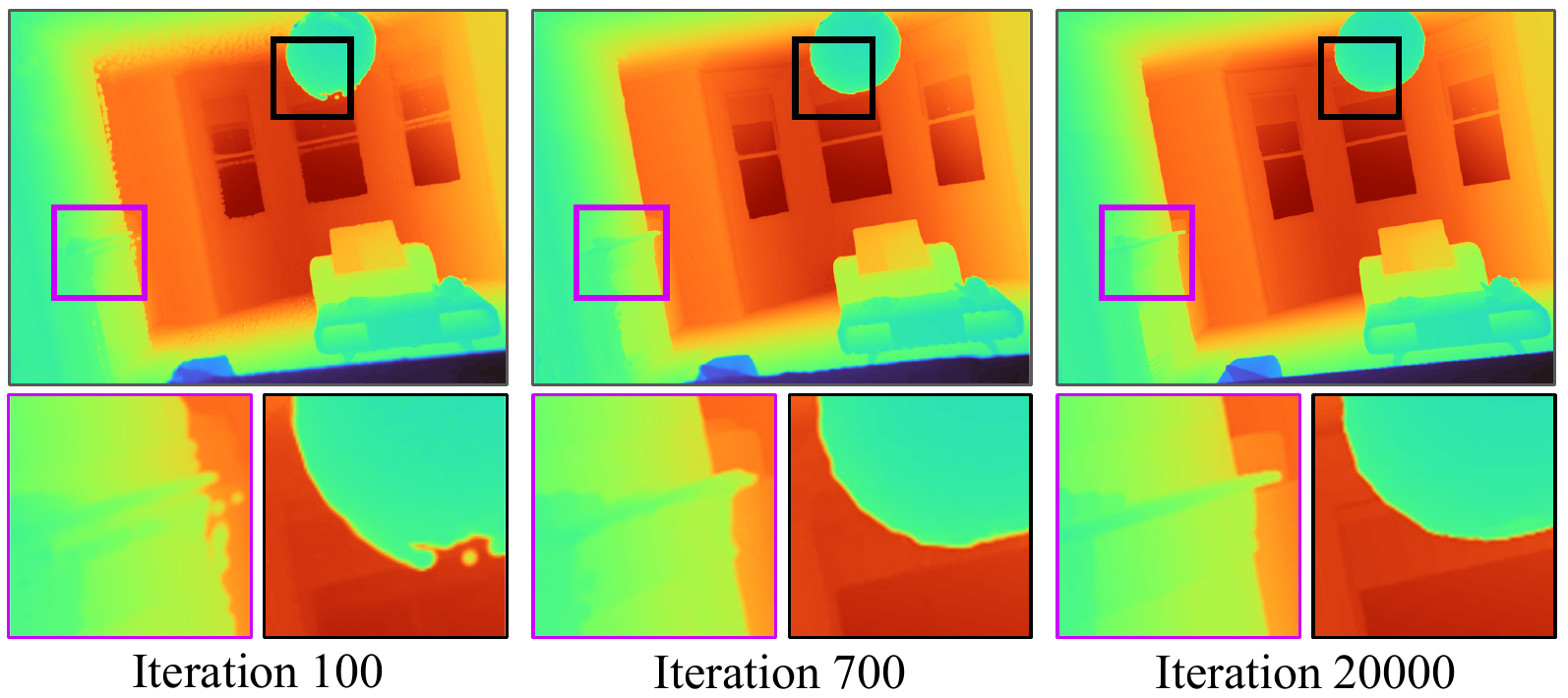}
  \vspace{-6mm}
  \caption{
The refinement process corrects misaligned 3D Gaussians by updating camera poses and geometry, with effectiveness demonstrated in Table~\ref{tab:ablation}.
  }
\label{fig:ba_effectiveness}
  \vspace{-3mm}
\end{figure}

This optimization aims to retain only Gaussians with an opacity of 1, pruning others to reduce ambiguity for the subsequent deblurring stage. We prune Gaussians with an opacity less than 0.8 and reset opacity with 0.5 again every certain amount of iteration.
As iterations progress, the rendered depth maps increasingly align with the input depth maps, indicating that all 3D Gaussian positions and camera poses are globally well-aligned (Figure~\ref{fig:ba_effectiveness}).

The final refinement loss is a weighted combination of the depth and opacity losses:
\begin{equation}
\label{eq:refinement_loss}
\mathcal{L}_\text{refinement }=  {\lambda _D}{\mathcal{L}_D} + {\lambda_o}{\mathcal{L}_\text{opacity}}.
\end{equation}

\subsection{Image Deblur}
\label{sec:deblur}
Motion blur caused by camera movement can generally be approximated as:
\begin{equation}
I(t) \approx B(t) = \frac{1}{k}\sum\limits_{i = 0}^k {{C_i}(t)}
\end{equation}
Here, $B(t)$ represents the motion-blurred image, which we model as the average of $k$ sharp images $C_i(t)$ rendered from the 3D Gaussians at $k$ virtual camera poses interpolated between the camera poses at the start and end of the exposure time period ${\bm{\xi}} _t^s$ and $ {\bm{\xi}} _t^e$. This approximation allows $B(t)$ to closely match the input image $I(t)$.

Using the refined camera poses and point cloud from the previous step, we incorporate depth information to achieve geometrically accurate deblurring, following the approach in recent work~\cite{zhao2024badgaussians}.
Specifically, we minimize the difference between the input image $I(t)$ and the averaged rendered image $B(t)$, as shown below:
\begin{equation}
{\mathcal{L}_{B}} = \sum\limits_{t} {\sum\limits_i {{{\left\| { {{I_t}}(\mathbf{p}_i) - {B}_{t}(\mathbf{p}_i)} \right\|}_1}} }.
\end{equation}
The final deblur loss function integrates this alignment loss with additional terms for structural similarity and depth consistency:
\begin{equation}
\label{eq:deblur_loss}
\mathcal{L}_\text{deblur }=  {(1 - \lambda _B)}{\mathcal{L}_B} + {\lambda_B}{\mathcal{L_\text{D-SSIM}}} + {\lambda _D}{\mathcal{L}_D}.
\end{equation}
This combined loss helps ensure that the deblurred output aligns well with both the input image and depth structure, leading to sharper and more geometrically consistent results.

\begin{figure*}
\centering
    \vspace{-2mm}
    \includegraphics[width=\linewidth]{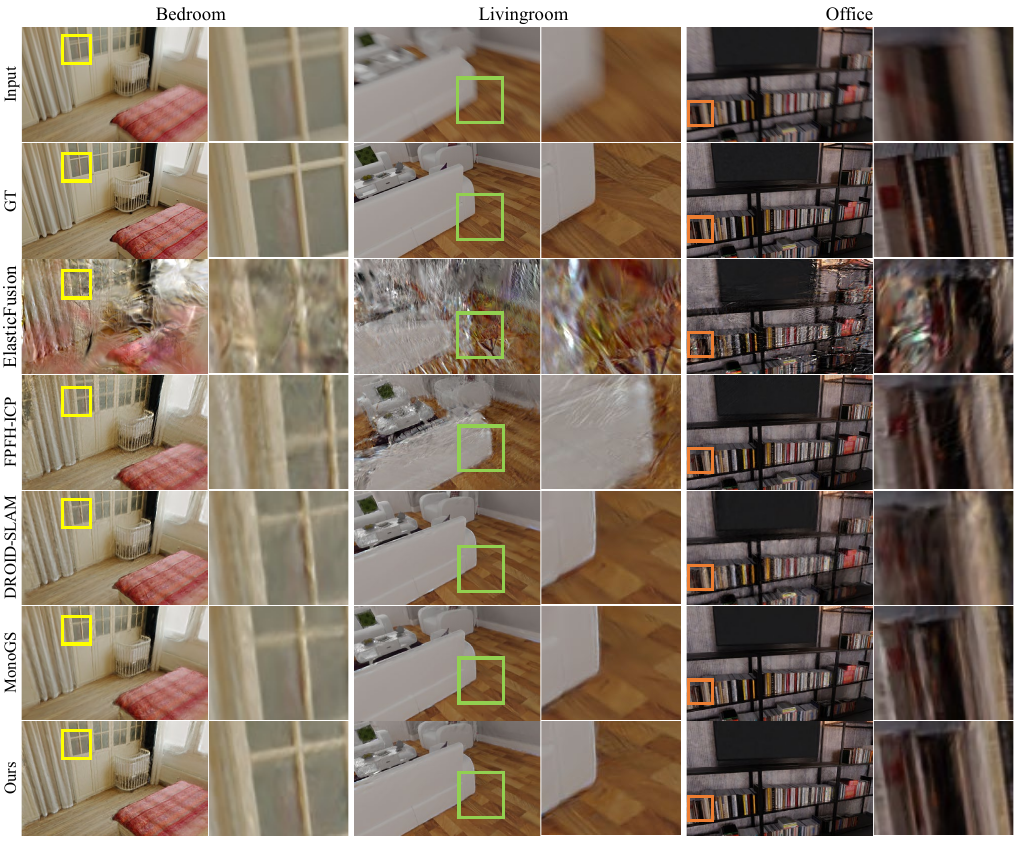}
    \vspace{-8mm}
    \caption{
    Qualitative comparison of deblurring performance from a novel view. Our method excels in restoring high-frequency details, whereas several methods, including COLMAP~\cite{schoenberger2016sfm}, fail under severe motion blur conditions (see Section~\ref{sec:evaluation_method}). The corresponding quantitative results are provided in Table~\ref{tab:deblur_accuracy}.
      }
    \label{fig:synthetic_result}
    \vspace{-2mm}
\end{figure*}

\begin{figure*}
\centering
    \vspace{-2mm}
    \includegraphics[width=\linewidth]{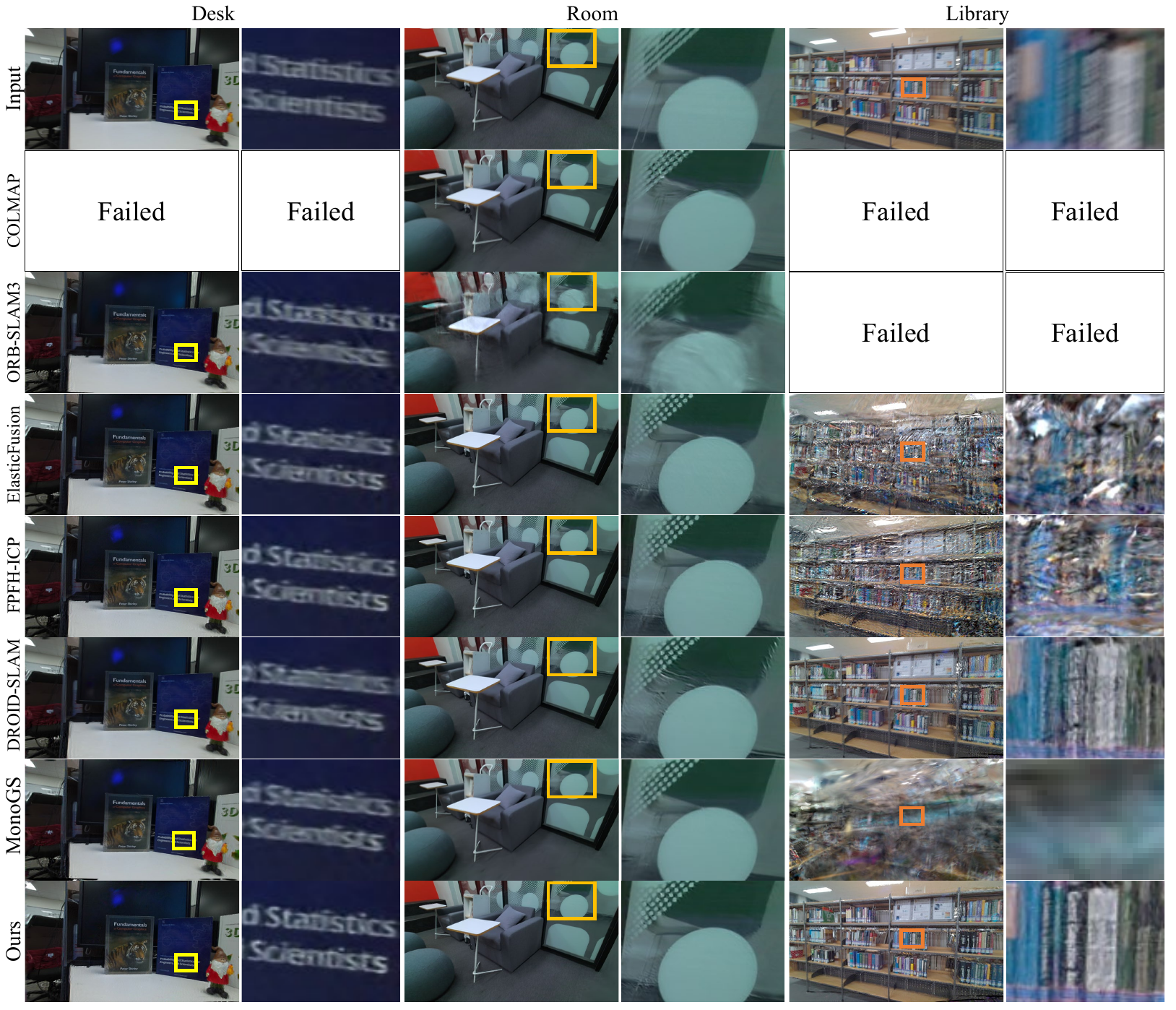}
    \vspace{-6mm}
    \caption{\label{fig:real_result}
    Qualitative comparison of deblurring performance from novel views on real scenes captured with an Azure Kinect camera. 
    Our method effectively restores high-frequency details, outperforming other methods even when processing RGB-D sequences with large spatial gaps between frames due to rapid camera motion and severe motion blur.}
    \vspace{-2mm}
\end{figure*}

\section{Results}
\label{sec:experiments}

\mparagraph{Dataset preparation}
We prepare both synthetic and real-world datasets to evaluate our method's performance.
In the absence of an RGB-D dataset with extreme motion blur, 
we generate a synthetic reference dataset by rendering directly from 3D models. 
We create RGB-D image sequences and record per-frame camera trajectories to evaluate our pose estimation and deblurring accuracy. 
The sequences, ranging from 50 to 150 frames, are captured along simulated trajectories that include rapid translations, 360-degree rotations, and human movement paths (using Blender’s Walk Navigation feature). 
The RGB and depth images have a resolution of $640\times480$, 
and motion blur is applied by enabling Blender's motion blur setting. 
The 3D models are sourced from the McGuire archive~\cite{McGuire2017Data} and the Blender archive~\cite{BlenderDemo}.

For real scene evaluation, 
we used an Azure Kinect RGB-D camera with a 33.3ms fixed exposure time, 5000K fixed white balance, 
NFOV-$2\times2$-binned depth capture mode, 
and $1280\times720$ RGB resolution. 
We used the provided intrinsic camera parameters to undistort the images and aligned depth maps to color images using the SDK’s transformation functions. 
Depth maps were eroded three times to remove outliers, which commonly occur along edges.

\mparagraph{Implementation details}
Our code is based on 3D Gaussian Splatting~\cite{kerbl20233d}, 
with data processing partially adapted from CF-3DGS~\cite{kerbl20233d}. 
For initial camera pose estimation, 
we use the Unimatch~\cite{xu2023unifying} optical flow network 
and the ICP~\cite{chen1992object} algorithm from Open3D as part of the optimization process.
For an input image sequence of $N$ frames, we prune Gaussians every 100$N$ iterations, 
reset opacity to 0.5 every 200$N$ iterations, and run a total of 600$N$ iterations (Section~\ref{sec:global_refinement}). 
The number of virtual views for deblurring is set to 15 for synthetic dataset evaluation and 17 for real-scene comparison. 
All experiments were conducted on an NVIDIA RTX A6000 GPU (48GB), with initial camera pose optimization taking 11 seconds per frame, 
0.1 seconds per iteration for pose and geometry refinement, and 0.26 seconds per iteration for deblurring at $640\times480$ resolution.

\subsection{Quantitative Evaluation}
\label{sec:evaluation_method}

We evaluate the camera pose accuracy and deblurring performance on three synthetic scenes from our dataset, each rendered with extreme motion blur. 
For pose accuracy, we compare our method against a variety of approaches capable of generating a 3D point cloud and camera trajectory: 
COLMAP~\cite{schoenberger2016sfm}, ICP with FPFH feature-based global alignment~\cite{rusu2009fast}, ORB-SLAM3~\cite{campos2021orb}, ElasticFusion~\cite{whelan2016elasticfusion}, and several RGB-D SLAM methods, including NICE-SLAM~\cite{zhu2022nice}, Point-SLAM~\cite{sandstrom2023point}, and DROID-SLAM~\cite{teed2021droid}. 
Additionally, we include recent Gaussian Splatting-based SLAM approaches such as MonoGS~\cite{matsuki2024gaussian} and SplaTAM~\cite{keetha2024splatam}. 
For each method, we calculate absolute trajectory error (ATE) and relative pose error (RPE) in meter and degree units, using the ground truth camera trajectories as references.

To evaluate deblurring accuracy, we modify a state-of-the-art Gaussian Splatting-based image restoration method~\cite{zhao2024badgaussians} to operate in a depth-aware manner using undistorted color and depth images. 
We select one out of every three images as a test view and render the corresponding novel view for comparison with the ground truth. 
For quantitative metrics, we calculate PSNR, SSIM, and LPIPS on the rendered images, 
as well as RMSE between the rendered and ground truth depth maps in the inverse depth domain ($m^{-1}$).

\definecolor{best_color}{HTML}{ade098}
\definecolor{second_best_color}{HTML}{f3e372}

  \begin{table*}[htp]
    \caption{
    Quantitative comparison of camera pose accuracy against ground truth poses. 
    We report scores only for methods that successfully provide both camera poses and a point cloud (see Section~\ref{sec:evaluation_method}). 
    Our method achieves the best scores on RPE metrics and shows ATE accuracy comparable to DROID-SLAM~\cite{teed2021droid}. 
    Each color highlights the \colorbox{best_color}{best} and \colorbox{second_best_color}{second best} results.
      }
    \label{tab:pose_accuracy}
    \vspace{-3mm}
    \resizebox{\textwidth}{!}{%
    \begin{tabular}{l|ccc|ccc|ccc} \hline
      & \multicolumn{3}{c|}{Bedroom}  & \multicolumn{3}{c|}{Livingroom}  & \multicolumn{3}{c}{Office}   \\ \hline
      & ATE $\downarrow$  & RPE (trans) $\downarrow$ & RPE (rot) $\downarrow$ & ATE $\downarrow$   & RPE (trans) $\downarrow$ & RPE (rot) $\downarrow$ & ATE $\downarrow$  & RPE (trans) $\downarrow$ & RPE (rot) $\downarrow$ \\ \hline
      ElasticFusion~\cite{whelan2016elasticfusion}   & 1.303 & 0.079 & 2.218 & 1.014 & 0.056 & 2.777 & 0.479 & 0.038 & 0.480    \\ 
      FPFH-ICP~\cite{rusu2009fast}                   & 2.760 & 0.118 & 2.310 & 2.028 & 0.092 & 1.213 & 0.139 & \cellcolor{second_best_color}0.009 & \cellcolor{second_best_color}0.167    \\ 
      DROID-SLAM~\cite{teed2021droid}                & \cellcolor{best_color}0.059 & 0.044 & 1.212 & \cellcolor{second_best_color}0.050 & 0.014 & 0.317 & \cellcolor{best_color}0.011 & 0.011 & 0.245    \\
      MonoGS~\cite{matsuki2024gaussian}              & 0.506 &\cellcolor{second_best_color} 0.041 & \cellcolor{second_best_color}0.812 & 0.135 & \cellcolor{second_best_color}0.009 & \cellcolor{second_best_color}0.124 & 0.045 & 0.011 & 0.180   \\ 
      Ours                                           & \cellcolor{second_best_color}0.102 &\cellcolor{best_color} 0.006 & \cellcolor{best_color}0.092 & \cellcolor{best_color}0.005 & \cellcolor{best_color}0.002 & \cellcolor{best_color}0.032 & \cellcolor{second_best_color}0.041 & \cellcolor{best_color}0.003 & \cellcolor{best_color}0.046    \\ \hline
      \end{tabular}%
    }
    \end{table*}

\begin{table*}[htp]
  \caption{
  Quantitative comparison of deblurring performance from novel views. 
  Our method demonstrates high accuracy across all metrics for both color and depth images.
  RMSE is calculated by comparing the rendered depth with GT depth in the inverse depth domain ($m^{-1}$).
   }
  \label{tab:deblur_accuracy}
  \vspace{-3mm}
  \resizebox{\textwidth}{!}{%
  \begin{tabular}{l|cccc|cccc|cccc} \hline
                & \multicolumn{4}{c|}{Bedroom}   & \multicolumn{4}{c|}{Livingroom}   & \multicolumn{4}{c}{Office}    \\ \hline
                & PSNR $\uparrow$  & SSIM $\uparrow$ & LPIPS $\downarrow$ & RMSE $\downarrow$  & PSNR $\uparrow$  & SSIM $\uparrow$ & LPIPS $\downarrow$ & RMSE $\downarrow$ & PSNR $\uparrow$  & SSIM $\uparrow$ & LPIPS $\downarrow$ & RMSE $\downarrow$ \\ \hline
  ElasticFusion~\cite{whelan2016elasticfusion}  & 18.766 & 0.586 & 0.466 & 1.249 & 13.799 & 0.448 & 0.554 & 0.146 & 19.052 & 0.533 & 0.401 & 0.079 \\
  FPFH-ICP~\cite{rusu2009fast}                  & 20.623 & 0.617 & 0.418 & 0.450 & 18.172 & 0.598 & 0.422 & 0.102 & 22.212 & 0.684 & 0.241 &   0.033 \\
  DROID-SLAM~\cite{teed2021droid}               &   21.375 &   0.666 &   0.302 &   0.013 &   21.222 &   0.735 &   0.250 &   0.035 & \cellcolor{best_color}25.680 & \cellcolor{second_best_color}0.769 & \cellcolor{best_color}0.144 & \cellcolor{best_color}0.013 \\
  MonoGS~\cite{matsuki2024gaussian}            & \cellcolor{second_best_color}23.728 & \cellcolor{second_best_color}0.717 & \cellcolor{second_best_color}0.285 & \cellcolor{second_best_color}0.009 & \cellcolor{second_best_color}24.414 & \cellcolor{second_best_color}0.822 & \cellcolor{second_best_color}0.188 & \cellcolor{second_best_color}0.027 &   24.194 &   0.725 &   0.184 & 0.184 \\
  Ours                                          & \cellcolor{best_color}26.745 & \cellcolor{best_color}0.824 & \cellcolor{best_color}0.206 & \cellcolor{best_color}0.005 & \cellcolor{best_color}27.650 & \cellcolor{best_color}0.900 & \cellcolor{best_color}0.150 & \cellcolor{best_color}0.010 & \cellcolor{second_best_color}25.649 & \cellcolor{best_color}0.791 & \cellcolor{second_best_color}0.160 & \cellcolor{second_best_color}0.014 \\ \hline
  \end{tabular}%
  }
  \vspace{-2mm}
  \end{table*}

\subsection{Ablation Study}
To validate the contributions of each component in our method, 
we perform an ablation study 
evaluating camera pose and deblurring accuracy by selectively omitting key elements. 
Specifically, we conduct experiments without fixing the scale of the Gaussians (Section~\ref{sec:global_refinement}),
without the pose and geometry refinement process (Equation~\ref{eq:refinement_loss}),
 and without depth loss (Equation~\ref{eq:deblur_loss}).

We assess the effectiveness of these components by performing deblurring 
based on results from Section~\ref{sec:initial-pose}—omitting refinement of camera poses and geometry—and by allowing the scale of the 3D Gaussians to be optimized freely. 
Table~\ref{tab:ablation} demonstrates that both the global refinement process and fixed Gaussian scale are essential for achieving high accuracy. 
Notably, during pose refinement, the algorithm reduces absolute pose error 
even at the expense of a slight increase in relative pose error, 
underscoring the importance of our refinement approach for stable, high-quality reconstruction.

\begin{table}[tp]
  \centering
   \caption{  
   Ablation study on the effects of the global refinement process and fixed scale of Gaussians. 
   Our refinement significantly enhances deblurring and depth accuracy while fixing the scale of the Gaussians leads to more accurate camera pose estimation.
   } 
   \label{tab:ablation}
   \vspace{-3mm}
  \resizebox{1.0\linewidth}{!}{%
  \begin{tabular}{l|cccc|ccc} \hline
    & PSNR  & SSIM  & LPIPS & RMSE  & ATE   & RPE(trans) & RPE(rot)  \\ \hline
  w/o scale fix       &   21.656 &   0.703 &   0.308 & 0.027  &   0.155  &   0.053 &   1.488  \\
  w/o refinement      & 22.223 & 0.734 & 0.238 & 0.027  & \cellcolor{second_best_color}0.123 & \cellcolor{best_color}0.007 & \cellcolor{best_color}0.047 \\
  w/o depth loss  & \cellcolor{second_best_color}25.518  & \cellcolor{second_best_color}0.807 & \cellcolor{best_color}0.170 & \cellcolor{second_best_color}0.014 & \cellcolor{best_color}0.103 & \cellcolor{second_best_color}0.012 & \cellcolor{second_best_color}0.195 \\
  Scale fix + refinement                & \cellcolor{best_color}26.681 &\cellcolor{best_color} 0.838 & \cellcolor{second_best_color}0.172 & \cellcolor{best_color}0.010  & \cellcolor{best_color}0.103 & \cellcolor{second_best_color}0.012 & \cellcolor{second_best_color}0.195 \\ \hline
  \end{tabular}
  }
  \vspace{-4mm}
  \end{table}

Our method achieves superior RPE scores across all scenes, 
as shown in Table~\ref{tab:pose_accuracy}, 
outperforming other approaches in pose and geometry estimation under severe motion blur. 
While our ATE scores are comparable to those of DROID-SLAM, 
which uses a bundle adjustment (BA) module, 
ATE alone does not correlate strongly with deblurring quality. 
Our method consistently produces more accurate and denser point clouds, which directly contributes to improved deblurring performance, 
as evidenced by the metrics in Table~\ref{tab:deblur_accuracy}. 
These results confirm that our integrated approach to pose estimation and deblurring is more resilient to challenging conditions than other methods.

\subsection{Qualitative Evaluation}

Figures~\ref{fig:synthetic_result} and \ref{fig:real_result} present qualitative comparisons of our method's performance in synthetic and real scenes. 
The densification capability of Gaussian Splatting provides detailed scene representation even from a relatively sparse point cloud; 
however, our approach, with its initially dense and accurate point cloud generation, achieves notably superior deblurring performance, especially in restoring high-frequency details. 
Our method consistently demonstrates a significant advantage in visual fidelity and detail restoration across diverse conditions.
\vspace{-2mm}

\section{Conclusion}
\label{sec:conclusion}

We have introduced a robust method for 3D scene reconstruction from RGB-D image sequences that effectively addresses the challenges of extreme motion blur and low-light conditions. 
Our approach leverages optical flow from color images and a carefully designed geometric loss to achieve accurate global alignment, followed by local refinement of camera poses and geometry using ICP. 
By integrating these steps into a Gaussian Splatting pipeline, we further refine camera poses and 3D geometry by minimizing depth and opacity losses. 
Additionally, fixing the scale of the 3D Gaussians ensures that the depth information from the RGB-D input is fully utilized, allowing for precise camera pose estimation and improved deblurring performance.
Our method demonstrates superior performance on real and synthetic RGB-D scenes with significant motion blur, outperforming existing approaches.

\appendix
\vspace{-0.1cm}
\section*{Acknowledgements}
\vspace{-0.15cm}
\noindent Min H.~Kim acknowledges the Samsung Research Funding \& Incubation Center (SRFC-IT2402-02), the Korea NRF grant (RS-2024-00357548), the MSIT/IITP of Korea (RS-2022-00155620, RS-2024-00398830, 2022-0-00058, and 2017-0-00072), Microsoft Research Asia, and Samsung Electronics.

\clearpage
{
    \small
    \bibliographystyle{ieeenat_fullname}
    \bibliography{main}
}

\clearpage % Start a new page
\pagestyle{empty} % Remove page numbering and headers/footers
\includepdf[pages=-]{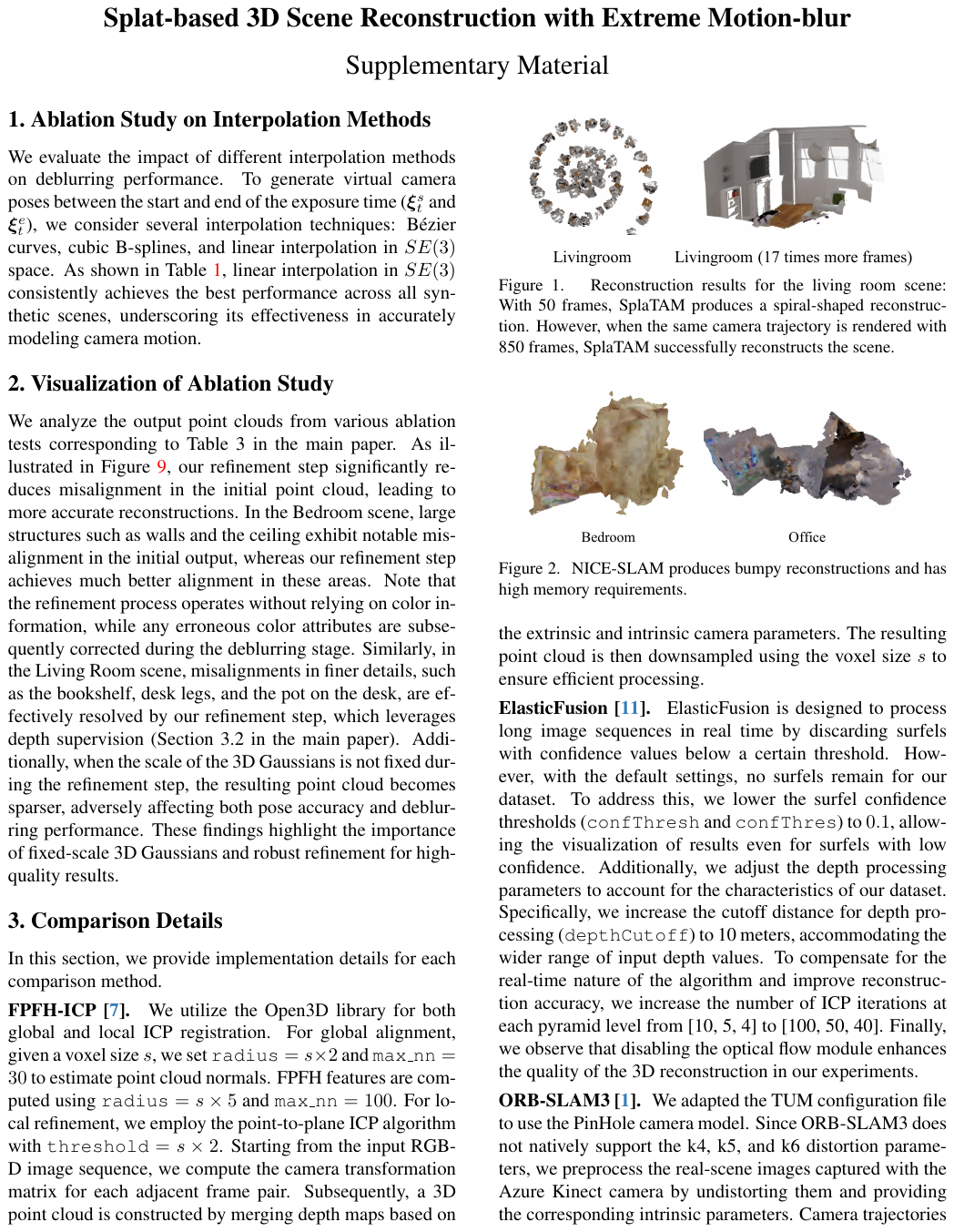}
\clearpage % Start a new page
\pagestyle{plain} % Restore page numbering and headers/footers

\end{document}